\documentclass[]{article} 

\usepackage{amssymb}
\usepackage{amsmath}
\usepackage{graphicx}
\usepackage{subfigure}
\usepackage{makeidx}
\usepackage{multicol}
\usepackage[export]{adjustbox}
\usepackage[justification=centering]{caption}
\usepackage{chngcntr}
\counterwithout{figure}{section}
\usepackage{float}
\usepackage{threeparttable}
\usepackage{graphicx}
\usepackage{xcolor} 

\newcommand{\blue}[1]{\textcolor{black}{#1}}

\usepackage[utf8]{inputenc}

\usepackage{fancyhdr}
\let\origtitle\title 
\renewcommand{\title}[1]{\lfoot{\textit{#1}}\origtitle{\textbf{#1}}}
\renewcommand{\sectionmark}[1]{\markboth {}{}}

\date{}

\title{Knowledge Distillation for Reservoir-based Classifier: Human Activity Recognition}

\begin{document}
\maketitle
\thispagestyle{fancy}
\centering

\author{Masaharu Kagiyama \footnote{kagiyama.masaharu410@mail.kyutech.jp}, Tsuyoshi Okita \footnote{tsuyoshi.okita@gmail.com}}\\
\thanks{$^1$$^2$Kyushu Institute of Technology}


\abstract{

\blue{This paper aims to develop an energy-efficient classifier for time-series data by introducing PatchEchoClassifier, a novel model that leverages a reservoir-based mechanism known as the Echo State Network (ESN). The model is designed for human activity recognition (HAR) using one-dimensional sensor signals and incorporates a tokenizer to extract patch-level representations. To train the model efficiently, we propose a knowledge distillation framework that transfers knowledge from a high-capacity MLP-Mixer teacher to the lightweight reservoir-based student model. Experimental evaluations on multiple HAR datasets demonstrate that our model achieves over 80 percent accuracy while significantly reducing computational cost. Notably, PatchEchoClassifier requires only about one-sixth of the floating point operations (FLOPS) compared to DeepConvLSTM, a widely used convolutional baseline. These results suggest that PatchEchoClassifier is a promising solution for real-time and energy-efficient human activity recognition in edge computing environments.}
}

\section{Introduction}
\label{section:Introduction}


Energy consumption in generative AI and large language models (LLMs) has become one of the urgent issues in deep learning. OpenAI reported that training the largest GPT-3 model \blue{consumed} approximately 1,300 MWh of electricity, which \blue{was} equivalent to the annual energy usage of about 130 U.S. homes \cite{OpenAIFewShot2020}. While there aren't many papers addressing the goal of low energy consumption in the context of LLMs, several reports indicate that some commercial LLMs have implemented this technology. This paper also aims to clarify the advancements in this area. Although \blue{our} research takes a different direction, there are studies that apply LLMs to the human activity recognition datasets we focus on. For example, recent studies~\cite{LLMHAR} and \cite{LLMHAR2} have explored the use of large language models (LLMs) for human activity recognition.


Currently, the technologies being explored in LLMs include quantization, pruning, and knowledge distillation. These methods help reduce model size and computational load, resulting in decreased energy consumption and power savings through hardware optimization. Additionally, the use of efficient algorithm designs, such as sparsification and early stopping, can effectively leverage computing resources and contribute to lower energy usage. These technologies can maintain model performance while reducing power consumption. Furthermore, multitask learning, distributed learning, and edge computing also play a role in energy savings. The adoption of dynamic model sizes and adaptive inference allows for efficient computation based on the input and task, minimizing unnecessary computational costs. Hyperparameter optimization and model reuse also contribute to reducing training time and energy consumption. These techniques enhance the energy efficiency of deep learning, particularly on edge devices and in mobile environments.


In this context, reservoir computing offers a unique perspective on power savings related to model compression and efficiency, as well as multitask learning, distributed learning, and edge computing. Regarding model compression and efficiency, this is primarily because the reservoir component remains frozen while only the output layer is trained. Optimizing the hardware reduces energy consumption and contributes to power savings. As for multitask learning, distributed learning, and edge computing, reservoir computing can be implemented using analog and optical devices, making it well-suited for real-time processing and time series prediction. This capability allows for processing on edge devices, which helps to reduce communication costs and energy consumption. Additionally, it facilitates distributed and lightweight calculations, demonstrating high performance with minimal resource usage.


Echo state networks (ESNs) are a type of reservoir computing that was first introduced in the early 2000s as a form of recurrent neural network \cite{Jaeger2004}. Their original goal was to address the challenges of training traditional RNNs, which often struggle with backpropagation over long sequences due to their recurrent connections. Since then, ESNs have been recognized for their shorter training times, reduced computational requirements, and lesser dependence on large training datasets compared to deep neural networks (DNNs). They also possess inherent simplicity and robustness, efficient training capabilities, minimal resource consumption, and a strong aptitude for approximating complex system dynamics. ESNs have been applied in biosignal classification to help reduce energy consumption \cite{Jiang2024}. In semiconductor research, power savings have also been targeted through the use of reservoir computing \cite{Yu2021}. In optical reservoir computing, both power savings and speed improvements have been emphasized \cite{Ludge2024}.


Liquid state machines (LSMs) are a type of reservoir computing and serve as a computational model particularly well-suited for time series data and real-time processing \cite{Maass2004}. The defining feature of LSMs is that the reservoir component changes dynamically like a "liquid," exhibiting complex and nonlinear responses to input. This capability arises from the use of spiking neurons in LSMs, in contrast to ESNs, which utilize continuous states. LSMs excel in processing temporal information and mimic biological neural activity by conveying information through discrete spikes, making them ideal for neuroscientific applications. Consequently, LSMs specialize in simulating the dynamic neural circuits of the brain and in time-dependent pattern recognition, serving a distinct purpose compared to ESNs.


Neuromorphic engineering \cite{Monroe2014} shares a similar purpose with reservoir computing but significantly differs in its hardware orientation. This field focuses on designing and developing hardware and systems that mimic the neural circuits of the brain, incorporating principles from neuroscience to replicate the behavior of neurons and synapses using electronic circuits. The ultimate goal is to create a system that is more energy-efficient and capable of superior parallel processing compared to conventional digital computers. Additionally, a substantial amount of research is being conducted to translate reservoir computing principles into hardware \cite{Yaremkevich2024, Cao2022}.


Next, we would like to discuss the distillation method as an approach for training such architectures. The distillation method involves transferring knowledge from a large-scale, high-performance model (the teacher model) in deep learning to a smaller, lighter model (the student model). The student model can conserve computational resources and enhance inference speed while maintaining accuracy that is close to that of the teacher model. This technique is particularly useful in resource-constrained environments, such as mobile and edge devices, and it also contributes to real-time processing and energy savings.

In image recognition, the teacher model is often a large-scale ResNet, while the student model is typically a lightweight MobileNet \cite{DeiT}. In object detection, knowledge is extracted from high-precision models like YOLO and SSD to develop lightweight models that operate in real time \cite{Wang2024}. DistilBERT is a language model created using the distillation method in natural language processing (NLP) \cite{Sanh2020}. DistilBERT distills knowledge from the BERT teacher model, resulting in a lightweight student model that improves computational efficiency and inference speed.


Generally, the model designated as the student model is often similar to the teacher model. However, in this study, the teacher model is a large language model (LLM) and the student model is a liquid state machine—two models with significantly different architectures. To the best of the author's knowledge, this form of distillation is rare.


The contributions of this paper are as follows:
\begin{itemize}
\item We propose a distillation model that transfers knowledge from an one dimensional MLP-Mixer to reservoir networks.
\begin{itemize}
\item The architectures of the teacher and student models are entirely different; specifically, the teacher is a vision-based MLP-Mixer, while the student is a reservoir network.
\end{itemize}
\item We deploy this distillation model to handle the one dimensional sensor signal data applied to human activity recognition task.
\end{itemize}


The structure of this paper is as follows: Section 2 discusses related research. Section 3 introduces our proposed model. Sections 4 and 5 describe the experimental setup. Section 6 presents the experimental results, while Section 7 offers discussions based on the results. Finally, Section 8 concludes the paper.\\

\section{Related Research}

\subsection{Energy-Efficient Networks}  
Research on achieving energy-efficient networks has been extensively conducted in the field of deep learning models. For instance, AutoSNN~\cite{autosnn} focuses on improving energy efficiency in Spiking Neural Networks, which model the behavior of biological neurons. This study emphasizes architectural design by eliminating energy-inefficient layers and introducing max-pooling layers for downsampling. As a result, it achieves high energy efficiency while maintaining accuracy comparable to other methods. 

\subsection{Knowledge Distillation}  
Knowledge distillation is a model compression technique that transfers knowledge from a teacher model with a large number of parameters to a student model with fewer parameters. Here, we discuss recent research related to distillation.\\  

Touvron et al.~\cite{DeiT} proposed a data-efficient training method for Vision Transformers in image classification and introduced a method utilizing knowledge distillation. In their knowledge distillation approach, they incorporated a distillation token, allowing the model to learn from the teacher model’s output, thereby facilitating the utilization of the teacher model's knowledge throughout the entire model. As a result, their method demonstrated superior performance compared to conventional distillation techniques. In our study, we applied this method to one-dimensional signals.

ZHOU et al.~\cite{ESNKD} introduced knowledge distillation into Echo State Networks (ESNs) to improve time-series prediction performance and reduce computational costs on edge devices. In their approach, they introduced an assistant network in addition to the teacher and student models to guide the learning process. As a result, their method achieved better performance and improved efficiency on edge devices compared to conventional ESNs and other optimized ESN methods. While their study shares a similar direction with ours, the distillation method differs.

Shiya et al.~\cite{Quant} utilized knowledge distillation to prevent performance degradation in quantized reservoir computing and introduced the "Teacher-Student Mutual Learning" method. They employed the pre-quantization reservoir as the teacher model and the post-quantization reservoir as the student model, leveraging both the output distribution and intermediate feature maps for distillation. As a result, the quantized model reduced resource consumption on FPGA while mitigating accuracy degradation. While this study shares the objective of resource reduction with ours, we do not incorporate quantization and focus on HAR as the target domain.

Qin et al.~\cite{hilbert} proposed a cross-dimensional distillation method called Hilbert Distillation, demonstrating superior performance compared to existing approaches. This study utilizes Hilbert curves to map 3D feature maps into 1D representations, enabling efficient knowledge transfer to 2D models.\\  

Sun et al.~\cite{logit} introduced Logit Standardization, a technique where the teacher and student models do not share temperature values. By applying this method to multiple distillation techniques, the authors improved image classification performance. The approach standardizes logits of both teacher and student models using z-score normalization as a preprocessing step, allowing the student model to focus on learning the relationships between logits from the teacher model.\\  

Park et al.~\cite{relational} proposed Relational Knowledge Distillation, which considers the structural relationships between data points. This method demonstrated superior performance over the teacher model in image classification and metric learning tasks. By combining distance-wise loss and angle-wise loss, the student model learns higher-order structural information instead of just individual outputs.\\  

Zhao et al.~\cite{decoupled} introduced Decoupled Knowledge Distillation, which independently handles the distillation of target and non-target class knowledge. This approach improves the trade-off between training efficiency and distillation performance while achieving results comparable to or better than existing methods in image classification tasks. By incorporating unique hyperparameters for target and non-target class knowledge, the method adjusts weights according to importance, ensuring non-target knowledge does not compromise the predictive accuracy of the teacher model.\\ 

\subsection{Reservoir Computing}
Reservoir computing has recently gained attention as a machine learning framework suitable for processing time-series data with high energy efficiency. One of the main advantages of reservoir computing is its fast training speed. Since the internal states of the reservoir remain fixed, training is limited to adjusting the output layer weights, significantly reducing training time. Several studies have applied this approach to HAR tasks.

Samip et al.~\cite{HARreservoir} proposed an HAR model utilizing reservoir computing with spiking neural networks. Their implementation on Intel's Loihi2 chip enabled low-power, high-efficiency time series processing.

Enrico et al.~\cite{highspeed} leveraged reservoir computing to efficiently classify HAR tasks using photonic RC, achieving high-speed and low-power video processing. Their photonic RC employs an optical delay feedback loop.







\section{Our Proposed Model}
\blue{In this section, based on the literature reviewed above, we now introduce our proposed model, which combines an MLP-Mixer-based distillation approach with a reservoir-based classifier designed for energy-efficient HAR.} We first present an inference mechanism using ESNs, and then discuss how to train such an inference mechanism of ESNs. In the second half, we use the distillation mechanism for training purposes.\\


\begin{figure*}[htbp]
  \begin{center}
\includegraphics[width=12cm]{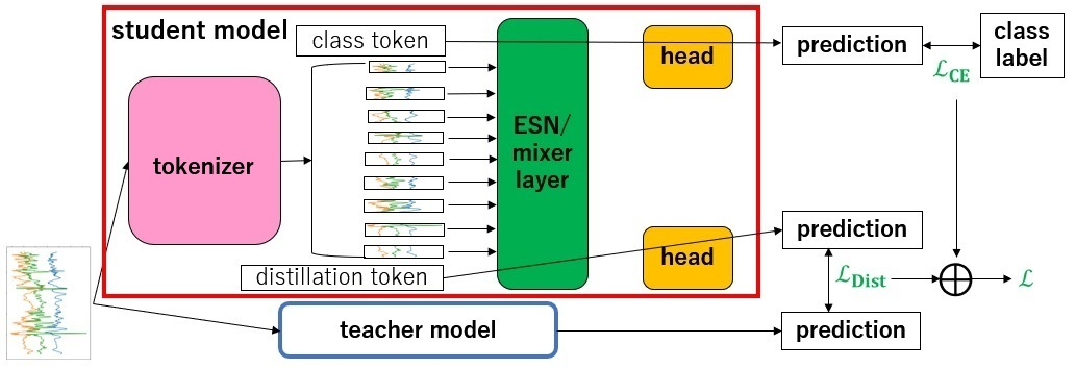}
  \caption{An overview of the proposed distillation method, Mixer-Echo State Signal Distillation. The input signals are fed into both the student and teacher models. Within the student model, after passing through the tokenizer layer, class and distillation tokens are added to the token set. These two tokens are then fed into separate heads to obtain the outputs. The output from the class token is used to compute the classification loss $L_{CE}$ along with the ground truth labels of the sensor data. The output from the distillation token is used to compute the distillation loss $L_{Dist}$ in conjunction with the output from the teacher model. The loss function of the proposed method is defined as a linear combination $L$ of $L_{CE}$ and $L_{Dist}$, and learning is performed to minimize this function.}
  \label{fig:dist}
  \end{center}
\end{figure*}

\subsection{Mixer-Echo State Signal Distillation: Distillation Mechanism}

In this experiment, we aim to reduce power consumption while maintaining a certain level of accuracy by distilling knowledge from MLP-Mixer like model, which have many parameters and achieve high accuracy. To achieve this, we adapted the distillation method of DeiT~\cite{DeiT}, originally designed for image models like ViT, to be adapted for one-dimensional signals and modified it to enable distillation into PatchEchoClassifier.\\


The distillation process in this experiment uses labeled data, along with both the teacher model and the student model. A summary of the distillation process is shown in Figure \ref{fig:dist}.\\



In the distillation method used here, after the student model's tokenizer process, we add a class token and a distillation token. These tokens are input into the layers following tokenizer, and their outputs are used in the distillation loss function. The process is explained mathematically in the following way.\\


We input the original signal $x$ and label $y$ into both the teacher and student models. After tokenizing in the student model, we obtain tokens $x^s_i (i=1,2,...)$. We then concatenate the class token $x_{cl}$ and the distillation token $x_{dist}$. This token sequence is fed into the subsequent layers, and we obtain the outputs of the class token $Z^s_{cl}$ and distillation token $Z^s_{dist}$. Additionally, the output $Z^t$ is obtained by feeding the signal $x$ into the teacher model.\\



The loss function in this study consists of the classification loss $L_{CE}$ and the distillation loss $L_{Dist}$. Each component is described as follows. For the classification loss, cross-entropy loss with label smoothing is employed. Label smoothing cross entropy is used to prevent overfitting by applying label smoothing to the cross-entropy loss, helping the student model's output distribution align more closely with the correct labels.\\


\begin{equation}
    \label{LCE}
    \begin{split}
    L_{CE} (Z^s_{cl}, y)= & - (1-\epsilon)\log (\mbox{softmax} (Z^{s (y=y)}_{cl})) \\ & -\epsilon  \dfrac{1}{N} \sum_{i=1}^N \log (\mbox{softmax} (Z^{s (y=y_i)}_{cl})
    \end{split}
\end{equation}


In Equation (\ref{LCE}), $\epsilon$ represents the Label Smoothing parameter, and $Z^{s (y=y)}_{cl}$ represents the probability of the correct label $y$ in the class token output of the student model. For the distillation loss, this study primarily employs KL divergence and JS divergence. KL divergence is a function used to measure the similarity between probability distributions and is used to bring the output distribution of the student model closer to that of the teacher model. In Equation (\ref{LDist}), $N (Z^s_{dist})$ refers to the number of elements in $Z^s_{dist}$, and $T$ is the temperature parameter.\\



\begin{equation}
    \label{LDist}
    \begin{split}
    & L_{Dist} (Z^s_{dist}, Z^t)= \\ & \dfrac{T^2}{N (Z^s_{dist})}\sum_{i=1}^N Z^{s (y=y_i)}_{dist}\log\dfrac{\mbox{softmax} (Z^{s (y=y_i)}_{dist}/T)}{\mbox{softmax} (Z^{t (y=y_i)}/T)}
    \end{split}
\end{equation}



JS divergence, like KL divergence, is a function used to measure the similarity between probability distributions; however, it is a symmetric function. This is shown in Equation (\ref{LDistJS}).\\

\begin{equation}
    \label{LDistJS}
    \begin{split}
    & L_{Dist} (Z^s_{dist}, Z^t)= \\ & \dfrac{1}{2}\sum_{i=1}^N Z^{s (y=y_i)}_{dist}\log\dfrac{\mbox{softmax} (Z^{s (y=y_i)}_{dist})}{\dfrac{1}{2} (\mbox{softmax} (Z^{s (y=y_i)}_{dist})+\mbox{softmax} (Z^{t (y=y_i)}))}\\
    & +
    \dfrac{1}{2}\sum_{i=1}^N Z^{t (y=y_i)}\log\dfrac{\mbox{softmax} (Z^{t (y=y_i)})}{\dfrac{1}{2} (\mbox{softmax} (Z^{s (y=y_i)}_{dist})+\mbox{softmax} (Z^{t (y=y_i)}))}
    \end{split}
\end{equation}


The overall distillation loss function is expressed as a linear combination of these two loss functions. As the distillation loss $L_{Dist}$, either KL divergence or JS divergence is adopted. In Equation (\ref{L}), $\alpha$ is the weight parameter for each loss function. This loss function corresponds to the distillation method referred to as soft distillation in the original DeiT.\\
\begin{equation}
\label{L}
L = (1-\alpha) L_{CE} + \alpha L_{Dist}
\end{equation}




In this experiment, the distillation process brings the student model's output closer to the correct labels at the same time as aligning it with the output of the teacher model's classification layer, thereby training the student model. The entire distillation process as described above is shown in Figure \ref{fig:dist}.\\


\subsection{PatchEchoClassifier: Inference Model}
\begin{figure*}[htbp]
  \begin{center}
 \includegraphics[width=13cm]{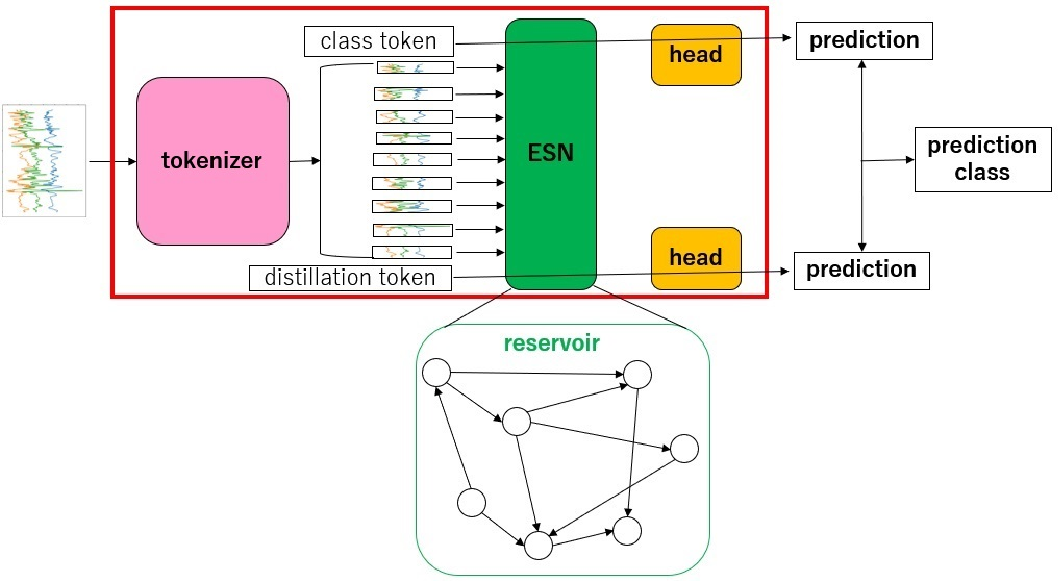}
 \caption{The architecture of PatchEchoClassifier. As shown in Figure \ref{fig:dist}, the tokenizer layer divides continuous signals with multiple channels into fixed intervals. Then, the sensor data, along with the class token and distillation token used in this method, are input into the Echo State Network (ESN). The ESN consists of a reservoir layer where matrix operations are performed. The graph of the reservoir layer represents the weight matrix of the ESN, and the number of nodes is determined by the size of the ESN. The two types of tokens used for distillation are then fed into their respective heads to obtain the outputs. During distillation training, the outputs from each head are used with separate loss functions. However, during inference, the class prediction is obtained by averaging the outputs from each head.}

  \label{fig:PatchEchoClassifier}
  \end{center}
\end{figure*}


We propose a model called PatchEchoClassifier, where the encoder part of a vision transformer~\cite{ViT} is replaced with an ESN, a type of reservoir, as shown in Figure \ref{fig:PatchEchoClassifier}, which shows the structure of the ESN. Let us denote $x$ the original signal and $y$ the label of $x$. When the ESN receives time-series input $x (=\{ x_1,\ldots,x_t \})$, the \blue{reservoir initialized} with random weights temporarily holds the data and combines it with the previous states to produce an output. One advantage of the ESN is that the reservoir weights are not updated during training, allowing for faster learning.\\


Although it would have been sufficient to use a reservoir with only the ESN, prior research on MetaFormer~\cite{MetaFormer} suggests that the success of models like ViT and MLP-Mixer is not attributed to the token mixer itself, but rather to the overall architecture, where processing is performed by the token mixer after tokenizing, followed by output generation. In fact, the MetaFormer~\cite{MetaFormer} paper demonstrates that a model with the token mixer replaced by pooling layers still achieves higher accuracy than ResMLP. Based on this, the proposed PatchEchoClassifier replaces the token mixer with an ESN.\\


Moreover, this modification in PatchEchoClassifier facilitates the application of DeiT~\cite{DeiT}, a distillation method that has achieved high accuracy in the field of image processing. As will be explained later, DeiT modifies tokens after tokenizer for use in distillation. This allows us to leverage a proven method from the image domain rather than starting from scratch when performing distillation on one-dimensional signals. \blue{Additionally, ESNs are particularly suitable for edge environments due to their fixed internal states and minimal training requirements, while our distillation mechanism bridges the performance gap with more powerful models.}\\


In the tokenizing phase, the signal data, which may consist of multiple channels, is split into patches by combining all channels at a fixed time point. Although it is possible to process each channel separately, combining them allows the model to learn the relationships between the channels, such as the 3-axis accelerometer data at the same time point, which is more desirable.\\



When the patch size is $p$, and the signal $x (=\{ x_1, \ldots, x_t \})$ is divided into patches, the $i$-th patch is represented as $X_i = \{x_{(i-1)p+1}, x_{(i-1)p+2},$ $ \ldots , x_{ip}\}$, where $X_i$ is a vector. In other words, the tokenizer layer segments the signal. The segments obtained through partitioning are generally referred to as tokens or patches. However, in this paper, we refer to them as patches. \blue{Specifically}, we obtain the vectors $X_i^{cls}$ and $X_i^{dist}$ by appending the class token $x_{cls(i)}$ and the distillation token $x_{dist(i)}$ \blue{at} the end, respectively.\\  

When $X_i^{cls}$ and $X_i^{dist}$ are fed into the ESN, the computation described in Equation (\ref{ESN_eq}) is performed, yielding the outputs $X^{ESN(cls)}$ and $X^{ESN(dist)}$. Here, $W_{\mbox{input}}$ is the weight matrix from the input layer to the reservoir layer, and $W_{\mbox{reservoir}}$ represents the weight matrix that defines the connections between neurons in the reservoir layer. The sizes of these two weight matrices are determined by the parameters of the ESN. Additionally, these weight matrices are initialized and fixed, meaning no learning occurs for these parameters. As a result, PatchEchoClassifier involves fewer parameters to learn, leading to reduced energy consumption, as demonstrated in the subsequent experiments.\\

\begin{equation}
    \label{ESN_eq}
    X^{ESN}_i=  \mbox{tanh} (X^{ESN}_{i-1} W_{\mbox{\blue{reservoir}}} + X_i W_{\mbox{input}}^T)
\end{equation}



Then, for each of the two heads, a linear function $y = WX^{ESN} + b$ is computed, yielding two outputs. At this time, the output of the head, which receives $X^{ESN}_{cls}$ generated by the ESN with the class token and all signal patches as input, is used for the classification loss $L_{CE}$. On the other hand, the output of the head, which receives $X^{ESN}_{dist}$ generated by the ESN with the distillation token and all signal patches as input, is used for $L_{DIST}$.\\  

During distillation training, these outputs are used in the calculation of the loss function as shown in Figure \ref{fig:dist}. However, during inference, as shown in Equation (\ref{inf_eq}), the class prediction is obtained by averaging the outputs from the two heads and using the resulting vector. In Equation (\ref{inf_eq}), $y_{cls}$ represents the output obtained from the head for the class token, and $y_{dist}$ represents the output obtained from the head for the distillation token.\\

\begin{equation}
    \label{inf_eq}
    y = \mbox{softmax} (\dfrac{y_{cls}+y_{dist}}{2})
\end{equation}

\begin{figure*}[htbp]
  \begin{center}
 \includegraphics[width=13cm]{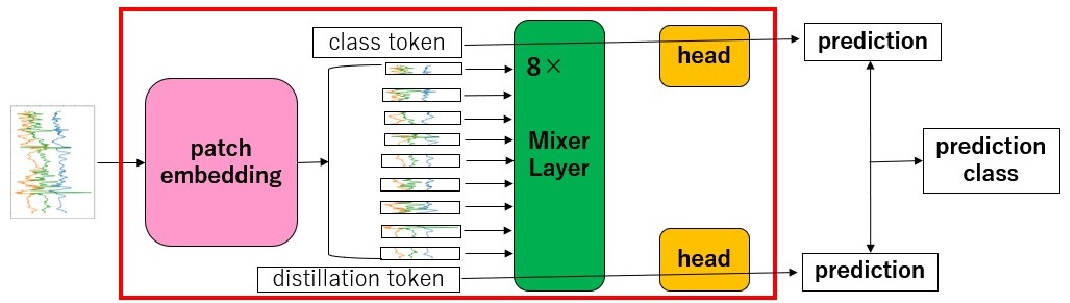}
 \caption{The architecture of PatchMixerClassifier. In the patch embedding layer, a continuous signal with multiple channels is segmented and position embedding is applied. Then, the sensor data, along with the class token and distillation token used in this method, are input into the Mixer Layers. In this experiment, the number of MixerLayers was set to 8. The two types of tokens used for distillation are then fed into their respective heads to obtain the outputs. During distillation training, the outputs from each head are used with separate loss functions. However, during inference, the class prediction is obtained by averaging the outputs from each head.}
  \label{fig:PatchMixerClassifier}
  \end{center}
\end{figure*}

The diagram shown in Figure \ref{fig:PatchEchoClassifier} provides an overview of the structure and processing flow of PatchEchoClassifier. The architecture of PatchMixerClassifier is shown in Figure \ref{fig:PatchMixerClassifier}. PatchMixerClassifier is a model designed for training using the proposed distillation method. It is an adaptation of MLP-Mixer~\cite{mlpmixer}, originally designed for processing 2D images, where the patch embedding layer has been modified to handle one-dimensional signals. In the patch embedding layer, position embeddings are added after partitioning, similar to the Vision Transformer. Additionally, similar to PatchEchoClassifier, two heads are prepared for distillation. During distillation, these heads are used for the loss functions $L_{CE}$ and $L_{Dist}$, respectively. During inference, the average of the prediction distributions from the two heads is used.

\section{Experimental Data}

The data used mainly in this study is from the SHL Challenge 2023 and 2024 ~\cite{SHL}. Additionally, four datasets were used for comparison experiments: the ADL dataset~\cite{adl}, PAMAP2 dataset~\cite{pamap2}, RealWorld dataset~\cite{realworld}, and WISDM dataset~\cite{WISDM}. These data include signals from sensors such as a 3-axis accelerometer and corresponding activity recognition labels. In this study, we used data from the 3-axis accelerometer for distillation. These datasets consist of data for human activity recognition.\\


To prevent overfitting with the limited sensor data, we applied data augmentation to the target signals. The data augmentation methods used are described below. First, we resampled the signal data to enable tokenizing. Then, we added jitter to the data based on a normal distribution.\\


Additionally, \blue{for} comparison, we adapted existing models for one-dimensional signal data. We created the 1DMLP-Mixer, based on MLP-Mixer~\cite{mlpmixer}, as teacher model. For the student models, we created PatchMixerClassifier and PatchEchoClassifier, which include token addition after \blue{tokenizer} and a classifier for each token to facilitate distillation. In addition, DeepConvLSTM~\cite{deepconvLSTM} was used as a baseline model for comparison. This model is designed for HAR and has a relatively small number of trainable parameters. However, unlike our proposed approach, this model was trained using a standard training process without distillation.\\

\section{Experimental Setup}


The distillation was carried out over 100 epochs, and the student model from the epoch with the highest accuracy on the validation data was selected. The parameters for each model were set as follows:\\


\begin{itemize}
    \item 1DMLP-Mixer (teacher) : patch size = 16, patch embedding layers dimension = 768, num of mixer layers = 12
    \item PatchMixerClassifier (student) : patch size = 16, tokenizer dimension =512, num of mixer layers=8
    \item PatchEchoClassifier (student) : the number of parameters and patch size in the reservoir are shown in the results table.
    \item DeepConvLSTM (for comparison) : We used DeepConvLSTM, which has 64 convolutional filters and 128 hidden units in the LSTM, as well as $\mbox{DeepConvLSTM}_{0.50}$ and $\mbox{DeepConvLSTM}_{0.25}$, where each parameter is reduced to half and one-fourth, respectively. However, the number of LSTM layers was fixed at 2.

\end{itemize}


For the dataset used in this study, we utilized the triaxial accelerometer data from the SHL 2023 or 2024 dataset, which was collected from a sensor attached to the waist. The training data was measured by user 1, while the validation data was measured by users 2 and 3, including both signal data and their corresponding activity recognition labels. Notably, the test data does not include labels, making it unsuitable for classification tasks; thus, it was not used. 

The SHL2023 training and validation data were segmented with a window size of 500, and the median label within each window was assigned as the label. A subset of 2,874 frames from the validation data was used as held-out test data, while the remaining data were used for training. Among the training data, 176,088 frames were used for distillation training, and 19,566 frames were used as validation data for distillation. From the validation data, excluding the held-out frames, 22,985 frames were used for fine-tuning in downstream tasks, and 2,873 frames were used as validation data. In this process, no overlap between training and test data due to sliding windows, as pointed out in previous research \cite{datasplit}, occurred. The batch size was set to 64.\\


The learning rate for the loss function during training was managed by a cosineLR scheduler. The first 5 epochs were treated as a warmup phase, during which the learning rate was incrementally increased. Afterward, the learning rate was reduced according to cosine annealing.\\

\section{Experimental Results}

First, we conducted a verification experiment to compare the differences in FLOPS when PatchEchoClassifier, PatchMixerClassifier, and DeepConvLSTM processed the same toy data. The toy data was generated as random numbers following a standard normal distribution and was fed into the models with a batch size of 64, 3 channels, and a signal length of 496 to compute the FLOPS.\\


From the FLOPS comparison shown in Table \ref{flops}, PatchEchoClassifier has lower FLOPS than PatchMixerClassifier. Notably, PatchEcho
$\mbox{Classifier}^{1000}_{p=128}$, which maintains a certain level of performance, performs inference in less than 2\% of the time relative to PatchMixerClassifier, thereby reducing energy consumption. Furthermore, $\mbox{PatchEchoClassifier}^{1000}_{p=128}$ achieves less than half the FLOPS compared to the existing model Deep
$\mbox{ConvLSTM}_{0.50}$. \\

\renewcommand{\arraystretch}{1.5}
\begin{table}[htbp]
\centering 
\caption{The comparison results of FLOPS for PatchEchoClassifier, PatchMixerClassifier, and DeepConvLSTM are presented. In this table, $\mbox{PatchEchoClassifier}^{S}_{p=a}$ indicates that the patch size is $a$ and the number of parameters in the ESN is $S$.}

\vspace{5pt}
\label{flops}
    \begin{tabular}{cccc}
        \hline
         \hspace{0.5em} & model& \hspace{1em} & FLOPS \\ \hline
         & PatchMixerClassifier & & 39985479680\\
         & $\mbox{DeepConvLSTM}_{}$ & & 9105719296\\
         & $\mbox{DeepConvLSTM}_{0.50}$ & & 2315460608\\
         & $\mbox{DeepConvLSTM}_{0.25}$ & & 598380544\\
         & $\mbox{PatchEchoClassifier}^{1000}_{p=32}$ & & \underline{92446720}\\
         & $\mbox{PatchEchoClassifier}^{1000}_{p=64}$ & & 341549056\\
         & $\mbox{PatchEchoClassifier}^{1000}_{p=128}$ & & 1161797632\\
         & $\mbox{PatchEchoClassifier}^{4000}_{p=128}$ & & 1164869632\\
        \hline
    \end{tabular}
\end{table}
\renewcommand{\arraystretch}{1.0}


The comparison of the maximum heap usage when inputting toy data with the same configuration into the models and obtaining the outputs is shown in Table \ref{heap}. Note that this comparison was performed on \blue{CPU}. Upon comparison, we observe that PatchEchoClassifier generally uses less heap memory than PatchMixerClassifier. On the other hand, when comparing DeepConvLSTM and PatchEchoClassifier, DeepConvLSTM uses a smaller heap memory size. The $\mbox{PatchEchoClassifier}^{1000}_{p=128}$ model, \blue{which retains acceptable performance}, uses less than one-fifth of the heap memory required by PatchMixerClassifier, resulting in reduced energy consumption.\\

\renewcommand{\arraystretch}{1.5}
\begin{table}[htbp]
\caption{The comparison results of heap memory size for PatchEchoClassifier, PatchMixerClassifier, and DeepConvLSTM are presented. In this table, $\mbox{PatchEchoClassifier}^{S}_{p=a}$ indicates that the patch size is $a$ and the number of parameters in the ESN is $S$.}

\vspace{5pt}
\label{heap}
\begin{center}
    \begin{tabular}{ccc}
        \hline
         model & heap area size (MB) \\ \hline
        PatchMixerClassifier & 4870\\
        $\mbox{DeepConvLSTM}_{}$ & 725\\
        $\mbox{DeepConvLSTM}_{0.50}$ & \underline{532}\\
        $\mbox{DeepConvLSTM}_{0.25}$ & 550\\
        $\mbox{PatchEchoClassifier}^{1000}_{p=32}$ & 977\\
        $\mbox{PatchEchoClassifier}^{1000}_{p=64}$ & 942\\
        $\mbox{PatchEchoClassifier}^{1000}_{p=128}$ & 874\\
        $\mbox{PatchEchoClassifier}^{4000}_{p=128}$ & 2030\\
        \hline
    \end{tabular}
    \end{center}
\end{table}
\renewcommand{\arraystretch}{1.0}



The total size of the Python libraries used for distillation training and the models is approximately 1.3GB. Of this, about 76\% is occupied by the deep learning framework PyTorch and GPU-related CUDA libraries.\\  

The comparison of the footprint sizes for each model is shown in Table \ref{footprint}. Upon comparison, \blue{we} observe that PatchEchoClassifier generally has a smaller footprint size than PatchMixerClassifier. When comparing DeepConvLSTM and PatchEchoClassifier, the footprint of $\mbox{DeepConvLSTM}_{0.25}$ is approximately 1/60th smaller relative to $\mbox{PatchEchoClassifier}^{1000}_{p=128}$. The $\mbox{PatchEchoClassifier}^{1000}_{p=128}$ model, which retains acceptable performance, has a footprint size less than 40\% that of PatchMixerClassifier, leading to reduced energy consumption. \\

\renewcommand{\arraystretch}{1.5}
\begin{table}[htbp]
\centering 
\caption{Comparison of footprint sizes  for PatchEchoClassifier, PatchMixerClassifier, and DeepConvLSTM In this table, $\mbox{PatchEchoClassifier}^{S}_{p=a}$ indicates that the patch size is $a$ and the number of parameters in the ESN is $S$.}

\vspace{5pt}
\label{footprint}
    \begin{tabular}{ccccc}
        \hline
         \hspace{1em}&model& \hspace{0.7em} & size (MB) \\ \hline
         &PatchMixerClassifier & &12.7\\
         &$\mbox{DeepConvLSTM}_{}$ & & 1.19\\
        &$\mbox{DeepConvLSTM}_{0.50}$ & & 0.304\\
        &$\mbox{DeepConvLSTM}_{0.25}$ & & \underline{0.0817}\\
         &$\mbox{PatchEchoClassifier}^{1000}_{p=32}$ & &4.21\\
         &$\mbox{PatchEchoClassifier}^{1000}_{p=64}$ & &4.37\\
        &$\mbox{PatchEchoClassifier}^{1000}_{p=128}$ & &4.78\\
        &$\mbox{PatchEchoClassifier}^{4000}_{p=128}$ & &66.5\\
        \hline
    \end{tabular}
\end{table}
\renewcommand{\arraystretch}{1.0}


Next, we compare the experimental results of distillation using various human activity recognition datasets. In this experiment, four datasets—ADL, PAMAP2, RealWorld, and WISDM—are used to compare the experimental results of PatchEchoClassifier and PatchMixerClassifier. The results are shown in Table \ref{testscore2}.\\

\renewcommand{\arraystretch}{1.4}
\begin{table*}[htbp]
    \centering
    \caption{Distillation of PatchMixerClassifier and PatchEchoClassifier using four types of 1D signal datasets for activity recognition, followed by classification on test data. The teacher models in the experiments in this table are all 1DMLP-Mixer. Additionally, PatchEchoClassifier with a patch size of 128 and 1000 parameters was used. The distillation loss function used was KL divergence. In this table, all indicators except accuracy are macro-averages.}

    \vspace{5pt}
    \label{testscore2}
    \setlength{\tabcolsep}{4pt}
    \begin{tabular}{ccccc}
        \hline
        data set (model) & accuracy (\%) & precision (\%) & recall (\%) & f1-score (\%)\\ \hline
        ADL (PatchMixerClassifier) & \underline{99.1} & \underline{98.9} & \underline{99.0} & \underline{98.9}\\
        ADL (PatchEchoClassifier) & 70.1 & 74.2 & 55.2 & 53.4\\ \hline
        PAMAP2 (PatchMixerClassifier) & \underline{97.7} & \underline{97.5} & \underline{97.5} & \underline{97.5}\\
        PAMAP2 (PatchEchoClassifier) & 66.8 & 67.5 & 59.4 & 56.8\\ \hline
        RealWorld (PatchMixerClassifier) & \underline{97.4} & \underline{97.8} & \underline{97.8} & \underline{97.7}\\
        RealWorld (PatchEchoClassifier) & 76.6 & 80.5 & 77.4 & 77.5\\ \hline
        WISDM (PatchMixerClassifier) & \underline{93.4} & \underline{94.3} & \underline{93.4} & \underline{93.5}\\
        WISDM (PatchEchoClassifier) & 59.8 & 60.5 & 59.8 & 58.9\\
        \hline
    \end{tabular}
\end{table*}

\renewcommand{\arraystretch}{1.0}


In Table \ref{testscore2}, PatchMixerClassifier shows high performance across all datasets. However, the performance gap between PatchMixerClassifier and PatchEchoClassifier varies depending on the dataset. For example, in the WISDM dataset, the accuracy drops by approximately 33\%, while in the RealWorld dataset, the drop is about 21\%. Therefore, PatchEchoClassifier may be useful for specific datasets. PatchEchoClassifier exhibits relatively high performance on datasets that include structured activities, such as ADL and RealWorld. On the other hand, PAMAP2 and WISDM contain diverse movements and are more complex datasets. Since the ESN used in PatchEchoClassifier is well-suited for handling simple and structured data, this likely explains the performance differences observed across these datasets.\\


Next, using the proposed distillation method, \blue{we trained PatchMixerClassifier, PatchEchoClassifier, and DeepConvLSTM as student models using 1DMLP-Mixer as the teacher}. After fine-tuning the student models obtained through distillation, we evaluated them on the test data. The results are presented in Table \ref{testscore}.

\renewcommand{\arraystretch}{1.4}
\begin{table*}[htbp]
    \centering
    \caption{Using the SHL 2023 dataset, we performed distillation on PatchMixerClassifier and PatchEchoClassifier, followed by fine-tuning. The classification results on the test data after fine-tuning are presented. In the case of DeepConvLSTM, the evaluation results on the test data are presented after pretraining followed by fine-tuning. However, random flipping was not used as a data augmentation technique in this case. In this table, $\mbox{PatchEchoClassifier}^{S}_{p=a}$ indicates that the patch size is $a$ and the number of parameters in the ESN is $S$. In the table, the "loss" column indicates the type of distillation loss used: KL refers to KL divergence, and JS represents JS divergence. Additionally, all indicators except accuracy are macro-averages.}
    \vspace{5pt}
    \label{testscore}
    \setlength{\tabcolsep}{4pt}
    \begin{tabular}{cccccc}
        \hline
        loss & student Model & accuracy (\%) & precision (\%) & recall (\%) & f1-score (\%)\\ \hline
        -&$\mbox{DeepConvLSTM}_{}$ & 90.6 & 92.4 & 90.7 & 91.5\\
        -&$\mbox{DeepConvLSTM}_{0.50}$ & 87.4 & 88.7 & 88.0 & 88.3\\
        -&$\mbox{DeepConvLSTM}_{0.25}$ & 80.2 & 81.9 & 80.6 & 80.6\\
        KL & PatchMixerClassifier & \underline{94.6} & \underline{95.5} & 94.2 & \underline{94.8}\\
        JS & PatchMixerClassifier & 94.4 & 94.6 & \underline{94.4} & 94.5\\
        KL & $\mbox{PatchEchoClassifier}^{1000}_{p=32}$ &  82.7 & 83.1 & 81.9 & 82.3\\
        JS & $\mbox{PatchEchoClassifier}^{1000}_{p=32}$ & 80.9 & 82.9 & 78.4 & 80.2\\
        KL & $\mbox{PatchEchoClassifier}^{1000}_{p=64}$ & 85.2 & 86.6 & 84.5 & 85.4 \\
        JS & $\mbox{PatchEchoClassifier}^{1000}_{p=64}$ & 85.1 & 87.3 & 84.0 & 85.4\\
        KL & $\mbox{PatchEchoClassifier}_{p=128}^{1000}$ & 86.0 & 87.7 & 85.7 & 86.6\\
        JS & $\mbox{PatchEchoClassifier}^{1000}_{p=128}$ & 82.6 & 84.2 & 82.5 & 82.8\\
        KL & $\mbox{PatchEchoClassifier}^{4000}_{p=128}$ & 87.0 & 88.5 & 87.1 & 87.7 \\
        JS & $\mbox{PatchEchoClassifier}^{4000}_{p=128}$ & 88.0 & 89.8 & 87.5 & 88.5\\
        \hline
    \end{tabular}
\end{table*}

\renewcommand{\arraystretch}{1.0}


In Table \ref{testscore}, when PatchMixerClassifier was used as the student model, the accuracy was approximately 9\% higher relative to $\mbox{PatchEchoClassifier}_{p=128}^{1000}$. Additionally, comparing $\mbox{DeepConvLSTM}_{0.25}$ and $\mbox{PatchEchoClassifier}_{p=128}^{1000}$, the accuracy of $\mbox{PatchEchoClassifier}_{p=128}^{1000}$ was approximately 6\% higher.\\


Additionally, it was also observed that increasing the ESN size or the patch size led to improved accuracy. Here, we consider the performance differences depending on whether the loss function is based on KL divergence or JS divergence. For PatchEchoClassifier, which has a smaller number of parameters, the performance difference between KL and JS divergence is minimal. In general, models with fewer parameters struggle to capture the complex shape of the teacher distribution adequately. As a result, the impact of the differences in characteristics between KL divergence and JS divergence on the final outcome is reduced, leading to a smaller performance gap. On the other hand, for PatchEchoClassifier with a larger number of parameters, the performance superiority of KL or JS divergence varies depending on the case. The data distribution obtained from the SHL 2023 dataset is multimodal due to the inclusion of movement-related activities. Consequently, JS divergence attempts to capture multiple peaks, which can interact with the randomness of the ESN's internal states, leading to higher variability in performance.\\

\section{Comparison of Energy Efficiency}

This section discusses the energy efficiency of each model based on the experimental results. In this paper, we define the following score, Energy Efficiency Score (EES), and use it as an evaluation metric for energy efficiency. The EES is a comprehensive indicator of total energy consumption, where a smaller value indicates better energy efficiency. Furthermore, all metrics used in EES are logarithmically transformed using $\log(1+x)$ and then normalized using min-max normalization. In Equation (\ref{energy}), $\alpha, \beta,$ and $\gamma$ represent the weights of each component, which vary depending on the system's characteristics. Multiple combinations are considered for comparison.

\begin{equation}
\label{energy}
    \begin{split}
&\mbox{Energy Efficiency Score (EES)} 
\\&=\alpha*\mbox{FLOPS} +\beta*\mbox{heap size}+\gamma*\mbox{footprint size}
    \end{split}
\end{equation}

Additionally, to evaluate not only energy consumption but also performance simultaneously, we define the Accuracy-to-Energy Ratio (AER) and use it as a metric to assess the balance between performance and energy efficiency. A larger value of AER indicates a better balance between energy efficiency and performance. In Equation (\ref{AER}), $\epsilon$ is a constant to prevent division by zero, and we set $\epsilon=10^{-6}$.

\begin{equation}
\label{AER}
\mbox{Accuracy-to-Energy Ratio (AER)}=\dfrac{\mbox{accuracy}}{\mbox{EES}+\epsilon}
\end{equation}

The FLOP, heap size, and footprint size used to calculate EES are based on the results presented in the table from the previous section. Similarly, the accuracy used in the calculation of AER is selected from the results shown in Table\ref{testscore}, specifically choosing the highest accuracy among different loss function configurations.

Next, in Table \ref{energycompare}, we set specific values for $\alpha, \beta,$ and $\gamma$ based on the expected system characteristics and perform a comparison of the models. As conditions for the three coefficient parameters, we considered the following four scenarios. First, as a benchmark, we use the Balance type, which takes the average of the three indicators. Next, for IoT devices where memory usage is the highest priority, we introduce the Memory-saving type, which assigns a large weight to heap size. For edge AI which requires less power consumption, we define the Power-saving type, which places greater emphasis on FLOPS. Lastly, for mobile applications that need to minimize storage capacity, we consider the Storage-optimized type, which prioritizes the model’s footprint. The specific parameter values for each type are provided in the caption of the corresponding table.

\renewcommand{\arraystretch}{1.4}
\begin{table*}[htbp]
\centering 
\caption{The comparison results of the performance and energy efficiency balance metric, AER, for PatchEchoClassifier and PatchMixerClassifier are shown. In this table, $\mbox{PatchEchoClassifier}^{S}_{p=a}$ indicates that the patch size is $a$ and the number of parameters in the ESN is 
$S$. Additionally, the parameters for each type are set as follows.\\
Balanced:$(\alpha, \beta, \gamma)=(1/3,1/3,1/3)$, Memory-saving (IoT) : $(\alpha, \beta, \gamma)=(0.2,0.5,0.3)$, Power-saving (edge AI): $(\alpha, \beta, \gamma)=(0.7, 0.2, 0.1)$, Storage-optimized (mobile apps) : $(\alpha, \beta, \gamma)=(0.2, 0.2, 0.6)$}
\label{energycompare}
    \begin{tabular}{ccccc}
        \hline
         model& Balanced & Memory-saving & Power-saving & Storage-optimized\\ \hline
         PatchMixerClassifier & 1.09 & 1.07 & 0.98 & 1.23\\
        $\mbox{DeepConvLSTM}_{}$ & 2.55 & 3.33 & 1.58 & 3.22\\
        $\mbox{DeepConvLSTM}_{0.50}$ & 4.55 & 7.30 & 2.32 & 6.56\\
        $\mbox{DeepConvLSTM}_{0.25}$ & \underline{7.46} & \underline{11.6} & 3.67 & \underline{12.4}\\
        $\mbox{PatchEchoClassifier}^{1000}_{p=32}$ & 3.79 & 3.29 & \underline{8.90} & 2.92\\
        $\mbox{PatchEchoClassifier}^{1000}_{p=64}$ & 2.97 & 2.96 & 3.53 & 2.60\\
        $\mbox{PatchEchoClassifier}^{1000}_{p=128}$ & 2.47 & 2.71 & 2.27 & 2.32\\
        $\mbox{PatchEchoClassifier}^{4000}_{p=128}$ & 1.31 & 1.28 & 1.71 & 1.09\\
        \hline
    \end{tabular}
\end{table*}
\renewcommand{\arraystretch}{1.0}

\begin{figure}[htbp]
\centering
  \includegraphics[width=0.9\textwidth]{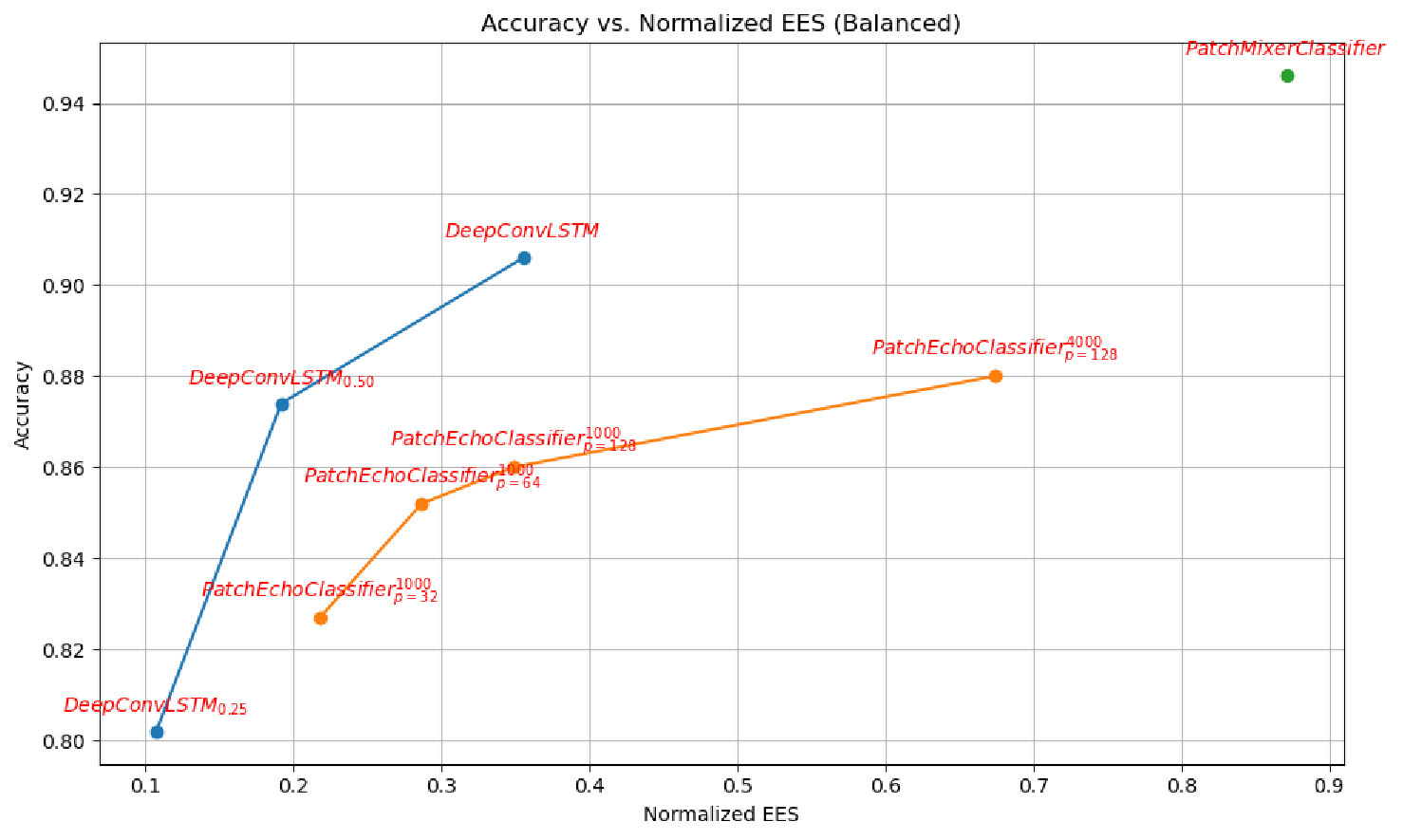}
  \caption{For each model, a line is plotted with accuracy on the vertical axis and the normalized value of the Energy Efficiency Score (EES) on the horizontal axis. Each point is labeled with the corresponding model name. The EES calculation is based on the balanced setting with $(\alpha, \beta, \gamma) = (1/3, 1/3, 1/3)$.}
  \label{fig:compare1}
\end{figure}

\begin{figure}[htbp]
  \centering
  \includegraphics[width=0.9\textwidth]{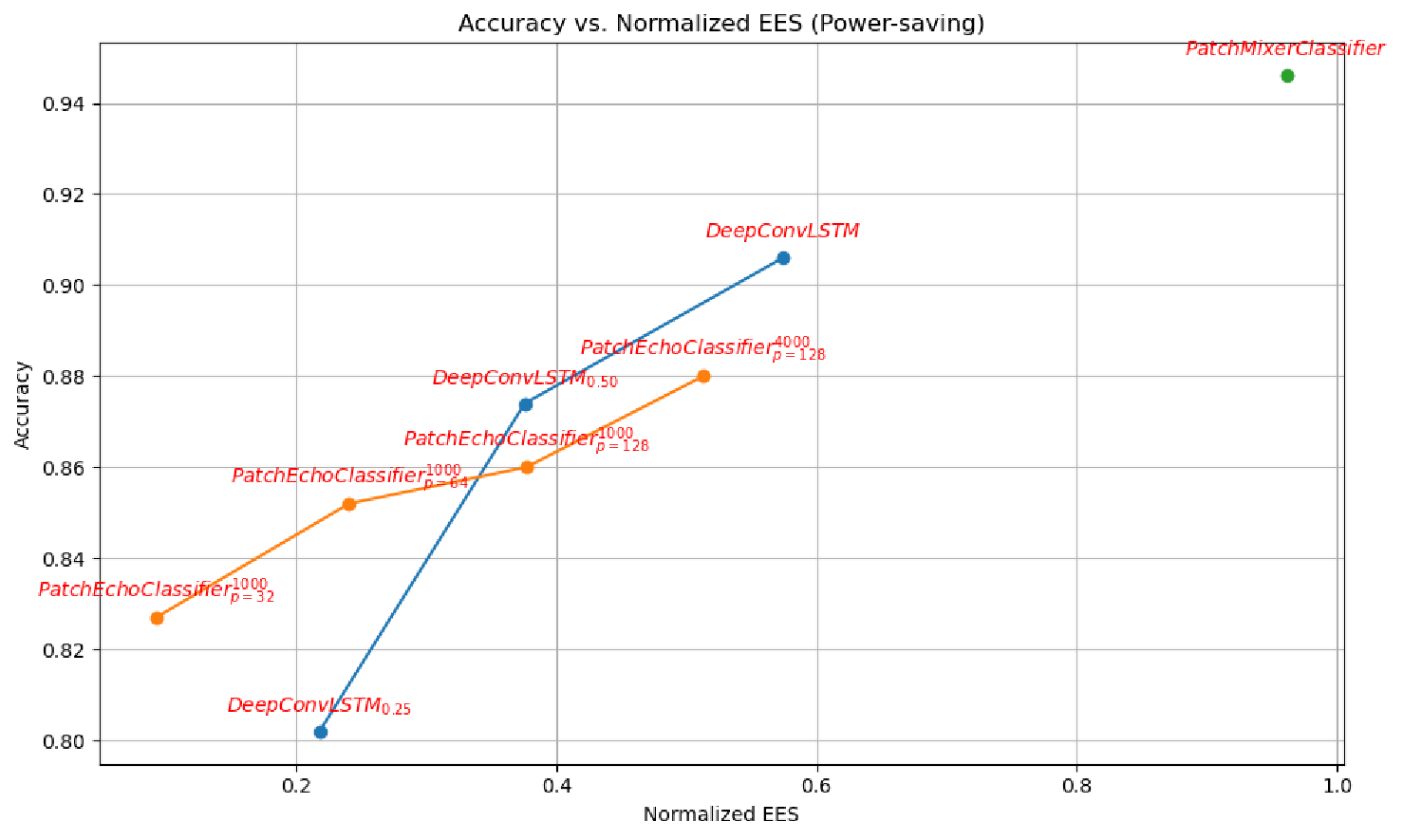}
  \caption{For each model, a line is plotted with accuracy on the vertical axis and the normalized value of the Energy Efficeincy Score (EES) on the horizontal axis. Each point is labeled with the corresponding model name. The EES calculation is based on the power-saving setting with $(\alpha, \beta, \gamma) = (0.7, 0.2, 0.1)$.}
  \label{fig:compare3}
  \end{figure}

As shown in \ref{energycompare}, except for the Power-saving type, $\mbox{DeepConvLSTM}_{0.25}$ achieves the highest AER in all cases, making it the model with the best balance between energy efficiency and performance. In contrast, for the Power-saving type, which assumes scenarios such as edge AI, PatchEchoClassifier with 1,000 ESN parameters and a patch size of 32 achieves the highest AER. However, since DeepConvLSTM performs worse than PatchEchoClassifier, a detailed comparison will be conducted in the following parts. Furthermore, in the comparison of PatchEchoClassifier, we observed that increasing the number of ESN parameters and patch size improves performance but reduces energy efficiency. In addition to Table \ref{energycompare}, a curve plotting accuracy against the Energy Consumption Index (ECI) is presented in Figure \ref{fig:compare1} and Figure \ref{fig:compare3}.

From Figure \ref{fig:compare1}, it can be observed that $\mbox{PatchEchoClassifier}_{p=64}^{1000}$ is an intermediate model between DeepConvLSTM and $\mbox{DeepConvLSTM}_{0.25}$ in terms of both performance and energy consumption. In other words, it consumes less energy than DeepConvLSTM while maintaining better performance than $\mbox{DeepConvLSTM}_{0.25}$. Additionally, a trade-off between performance and energy consumption can be observed, where higher-performing models tend to have greater energy consumption. Among them, PatchMixerClassifier exhibits particularly high energy consumption relative to its performance, indicating that it is an excessively large model. However, DeepConvLSTM remains a well-balanced model in terms of both energy consumption and performance.


From Figure \ref{fig:compare3}, when focusing on power consumption reduction, Patch
$\mbox{EchoClassifier}_{p=32}^{1000}$ and $\mbox{PatchEchoClassifier}_{p=64}^{1000}$ exhibit better performance than DeepConvLSTM with comparable energy consumption. However, for 
$\mbox{PatchEchoClassifier}_{p=128}^{1000}$, DeepConvLSTM outperforms in terms of accuracy. From this perspective, PatchEchoClassifier demonstrates superior performance compared to conventional CNN models while maintaining low power consumption.

Based on these comparisons, PatchEchoClassifier is not well-suited for IoT devices where memory efficiency is a priority. However, it appears to be a promising candidate for edge AI applications, where energy efficiency is crucial while maintaining high performance. Furthermore, comparing PatchMixerClassifier and PatchEchoClassifier, it was found that PatchEchoClassifier can reduce energy consumption more effectively.

\section{Discussion}

In this research, we initially intended to implement hard distillation, which achieved higher accuracy than soft distillation in the original DeiT paper. However, in our experiments with one-dimensional signal data, hard distillation did not result in significant loss reduction, and learning did not proceed as expected. In contrast, the soft distillation method from DeiT, which was employed in this study, allowed for consistent loss reduction and smooth training.\\



Additionally, in an experiment conducted to determine whether PatchEchoClassifier is suitable for edge deployment, it was found to outperform PatchMixerClassifier in terms of inference execution time, model footprint size, and heap memory usage during inference. Moreover, in the classification task of SHL 2023, although PatchEchoClassifier falls short of PatchMixerClassifier in terms of accuracy, it still achieves an accuracy in the 80\% range. Considering the model's energy consumption, this performance is deemed sufficient. On the other hand, the large size of the required Python libraries for the backbone poses a challenge for deployment on edge devices.\\






To achieve PatchEchoClassifier performance within a 10\% accuracy difference from the MLP-Mixer, we considered employing ensemble learning. However, simply applying ensemble learning did not yield good results. In an ensemble model where three parallel $\mbox{PatchEchoClassifier}^{4000}$ models were used, the accuracy was 58.9\blue{\%}, which was lower than the accuracy of a single $\mbox{PatchEchoClassifier}^{8000}$ model.

\section{Conclusions and Further Avenues}\label{結論}


We proposed two models: a new time-series model PatchEchoClassifier that incorporates tokenizer and a reservoir, and a tailored distillation method to achieve energy efficiency. PatchEchoClassifier achieved over 80\% accuracy while maintaining lower energy consumption compared to MLP-Mixer-based models. Additionally, the comparison of energy consumption and performance demonstrated that PatchEchoClassifier achieves higher efficiency and superior performance compared to the existing CNN model, DeepConvLSTM, particularly in scenarios where power consumption needs to be minimized. This highlights its potential in addressing the current issue of high power consumption in artificial intelligence.\\


For future work, we aim to improve the model structure and learning methods to achieve a reservoir network that reaches comparable accuracy to the MLP-Mixer. As for improving the distillation method, we are exploring modifications to how the tokens are incorporated into the loss function. Additionally, we seek to further reduce energy consumption and develop a model that outperforms existing methods. Furthermore, by combining other techniques, such as quantization, we aim to develop a more energy-efficient model.

\bibliographystyle{plain}
\bibliography{bibtex}

\begin{thebibliography}{10}

\bibitem{deepconvLSTM}
Marius Bock, Alexander Hölzemann, Michael Moeller, and Kristof Van~Laerhoven.
\newblock Improving deep learning for har with shallow lstms.
\newblock In {\em 2021 International Symposium on Wearable Computers}, UbiComp
  ’21. ACM, September 2021.

\bibitem{OpenAIFewShot2020}
Tom~B. Brown, Benjamin Mann, Nick Ryder, Melanie Subbiah, Jared Kaplan,
  Prafulla Dhariwal, Arvind Neelakantan, Pranav Shyam, Girish Sastry, Amanda
  Askell, Sandhini Agarwal, Ariel Herbert-Voss, Gretchen Krueger, Tom Henighan,
  Rewon Child, Aditya Ramesh, Daniel~M. Ziegler, Jeffrey Wu, Clemens Winter,
  Christopher Hesse, Mark Chen, Eric Sigler, Mateusz Litwin, Scott Gray,
  Benjamin Chess, Jack Clark, Christopher Berner, Sam McCandlish, Alec Radford,
  Ilya Sutskever, and Dario Amodei.
\newblock Language models are few-shot learners.
\newblock {\em arXiv preprint {arXiv:2005.14165}}, 2020.

\bibitem{adl}
Barbara Bruno, Fulvio Mastrogiovanni, and Antonio Sgorbissa.
\newblock {Dataset for ADL Recognition with Wrist-worn Accelerometer}.
\newblock UCI Machine Learning Repository, 2012.
\newblock {DOI}: https://doi.org/10.24432/C5PC99.

\bibitem{Cao2022}
Jie Cao, Xumeng Zhang, Hongfei Cheng, Jie Qiu, Xusheng Liu, Ming Wang, and
  Qi~Liu.
\newblock Emerging dynamic memristors for neuromorphic reservoir computing.
\newblock {\em Nanoscale(14)2}, pages 289--298, 2022.

\bibitem{LLMHAR2}
Ian Cleland, Luke Nugent, Federico Cruciani, and Chris Nugent.
\newblock Leveraging large language models for activity recognition in smart
  environments.
\newblock In {\em 2024 International Conference on Activity and Behavior
  Computing (ABC)}, pages 1--8, 2024.

\bibitem{ViT}
Alexey Dosovitskiy, Lucas Beyer, Alexander Kolesnikov, Dirk Weissenborn,
  Xiaohua Zhai, Thomas Unterthiner, Mostafa Dehghani, Matthias Minderer, Georg
  Heigold, Sylvain Gelly, Jakob Uszkoreit, and Neil Houlsby.
\newblock An image is worth 16x16 words: Transformers for image recognition at
  scale.
\newblock {\em arXiv preprint arXiv:2010.11929}, 2021.

\bibitem{SHL}
H.~Gjoreski, M.~Ciliberto, L.~Wang, F.J.O. Morales, S.~Mekki, S.~Valentin, and
  D.~Roggen.
\newblock The university of sussex-huawei locomotion and transportation dataset
  for multimodal analytics with mobile devices, 2018.

\bibitem{ESNKD}
ZHOU Jian, JIANG Yuwen, XU~Lijie, ZHAO Lu, and XIAO Fu.
\newblock Echo state network based on improved knowledge distillation for edge
  intelligence.
\newblock {\em Chinese Journal of Electronics}, 33(1):101--111, 2024.

\bibitem{Jiang2024}
Sai Jiang, Jinrui Sun, Mengjiao Pei, Lichao Peng, Qinyong Dai, Chaoran Wu,
  Jiahao Gu, Yanqin Yang, Jian Su, Ding Gu, Han Zhang, Huafei Guo, and Yun Li.
\newblock Energy-efficient reservoir computing based on solution-processed
  electrolyte/ferroelectric memcapacitive synapses for biosignal
  classification.
\newblock {\em J Phys Chem Lett:15(33):8501-8509}, 2024.

\bibitem{HARreservoir}
Samip Karki, Diego~Chavez Arana, Andrew Sornborger, and Francesco Caravelli.
\newblock Neuromorphic on-chip reservoir computing with spiking neural network
  architectures, 2024.

\bibitem{WISDM}
Jennifer~R. Kwapisz, Gary~M. Weiss, and Samuel Moore.
\newblock Activity recognition using cell phone accelerometers.
\newblock {\em SIGKDD Explor.}, 12:74--82, 2011.

\bibitem{Quant}
Shiya Liu, Lingjia Liu, and Yang Yi.
\newblock Quantized reservoir computing for spectrum sensing with knowledge
  distillation.
\newblock {\em IEEE Transactions on Cognitive and Developmental Systems},
  15(1):88--99, 2023.

\bibitem{Ludge2024}
Kathy Ludge.
\newblock Photonic reservoir computing for energy efficient and versatile
  machine learning application.
\newblock {\em Proceedings of the Royal Society of Victoria}, pages 1--3, 2024.

\bibitem{Maass2004}
Wolfgang Maass and Henry Markram.
\newblock On the computational power of recurrent circuits of spiking neurons.
\newblock {\em Journal of Computer and System Sciences, 69 (4): 593–616},
  2004.

\bibitem{Monroe2014}
Wolfgang Maass and Henry Markram.
\newblock Neuromorphic computing gets ready for the (really) big time.
\newblock {\em Communications of the ACM 57 (6)}, pages 13–--15, 2014.

\bibitem{autosnn}
Byunggook Na, Jisoo Mok, Seongsik Park, Dongjin Lee, Hyeokjun Choe, and Sungroh
  Yoon.
\newblock Autosnn: Towards energy-efficient spiking neural networks, 2022.

\bibitem{relational}
Wonpyo Park, Dongju Kim, Yan Lu, and Minsu Cho.
\newblock Relational knowledge distillation, 2019.

\bibitem{highspeed}
Enrico Picco, Piotr Antonik, and Serge Massar.
\newblock High speed human action recognition using a photonic reservoir
  computer.
\newblock {\em Neural Networks}, 165:662–675, August 2023.

\bibitem{hilbert}
Dian Qin, Haishuai Wang, Zhe Liu, Hongjia Xu, Sheng Zhou, and Jiajun Bu.
\newblock Hilbert distillation for cross-dimensionality networks, 2022.

\bibitem{pamap2}
Attila Reiss and Didier Stricker.
\newblock Introducing a new benchmarked dataset for activity monitoring.
\newblock {\em 2012 16th International Symposium on Wearable Computers}, pages
  108--109, 2012.

\bibitem{Sanh2020}
Victor Sanh, Lysandre Debut, Julien Chaumond, and Thomas Wolf.
\newblock Distilbert, a distilled version of bert: smaller, faster, cheaper and
  lighter.
\newblock {\em arXiv preprint arXiv:1910.01108}, 2020.

\bibitem{LLMHAR}
Milyun~Ni’ma Shoumi and Sozo Inoue.
\newblock Leveraging the large language model for activity recognition: A
  comprehensive review.
\newblock {\em International Journal of Activity and Behavior Computing},
  2024(2):1--27, 2024.

\bibitem{logit}
Shangquan Sun, Wenqi Ren, Jingzhi Li, Rui Wang, and Xiaochun Cao.
\newblock Logit standardization in knowledge distillation, 2024.

\bibitem{Jaeger2004}
Harnessing Nonlinearity: Predicting~Chaotic Systems and Saving~Energy
  in~Wireless~Communication.
\newblock Jaeger, herbert and haas, herald.
\newblock {\em Science. 304 (5667)}, pages 78–--80, 2004.

\bibitem{realworld}
Timo Sztyler and Heiner Stuckenschmidt.
\newblock On-body localization of wearable devices: An investigation of
  position-aware activity recognition.
\newblock In {\em 2016 IEEE International Conference on Pervasive Computing and
  Communications (PerCom)}, pages 1--9, 2016.

\bibitem{datasplit}
Andrés Tello, Victoria Degeler, and Alexander Lazovik.
\newblock Too good to be true: accuracy overestimation in (re)current practices
  for human activity recognition.
\newblock In {\em 2024 IEEE International Conference on Pervasive Computing and
  Communications Workshops and other Affiliated Events (PerCom Workshops)},
  page 511–517. IEEE, March 2024.

\bibitem{mlpmixer}
Ilya Tolstikhin, Neil Houlsby, Alexander Kolesnikov, Lucas Beyer, Xiaohua Zhai,
  Thomas Unterthiner, Jessica Yung, Daniel Keysers, Jakob Uszkoreit, Mario
  Lucic, and Alexey Dosovitskiy.
\newblock Mlp-mixer: An all-mlp architecture for vision.
\newblock {\em arXiv preprint arXiv:2105.01601}, 2021.

\bibitem{DeiT}
Hugo Touvron, Matthieu Cord, Matthijs Douze, Francisco Massa, Alexandre
  Sablayrolles, and Herve Jegou.
\newblock Training data-efficient image transformers \& distillation through
  attention.
\newblock In {\em International Conference on Machine Learning}, volume 139,
  pages 10347--10357, July 2021.

\bibitem{Wang2024}
Jiabao Wang, Yuming Chen, Zhaohui Zheng, Xiang Li, Ming-Ming Cheng, and Qibin
  Hou.
\newblock Crosskd: Cross-head knowledge distillation for object detection.
\newblock {\em arXiv preprint arXiv:2306.11369}, 2024.

\bibitem{Yaremkevich2024}
Dmytro~D. Yaremkevich, Alexey~V. Scherbakov, Luke~De Clerk, Serhii~M.
  Kukhtaruk, Achim Nadzeyka, Richard Campion, Andrew~W. Rushforth, Sergey
  Savel’ev, Alexander~G. Balanov, and Manfred Bayer.
\newblock On-chip phonon-magnon reservoir for neuromorphic computing.
\newblock {\em Nature Communications volume 14 (8296)}, 2023.

\bibitem{Yu2021}
Jie Yu, Yi~Li, Wenxuan Sun, Woyu Zhang, Zhaomeng Gao, and Danian Dong.
\newblock Energy efficient and robust reservoir computing system using
  ultrathin (3.5 nm) ferroelectric tunneling junctions for temporal data
  learning.
\newblock {\em IEEE 2021 Symposium on VLSI Technology}, pages 1--2, 2021.

\bibitem{MetaFormer}
Weihao Yu, Mi~Luo, Pan Zhou, Chenyang Si, Yichen Zhou, Xinchao Wang, Jiashi
  Feng, and Shuicheng Yan.
\newblock Metaformer is actually what you need for vision, 2022.

\bibitem{decoupled}
Borui Zhao, Quan Cui, Renjie Song, Yiyu Qiu, and Jiajun Liang.
\newblock Decoupled knowledge distillation, 2022.

\end{thebibliography}

\end{document}